# COVID-19 Pandemic: Identifying Key Issues using Social Media and Natural Language Processing

Oladapo Oyebode, Chinenye Ndulue, Dinesh Mulchandani, Banuchitra Suruliraj, Ashfaq Adib, Fidelia Anulika Orji, Evangelos Milios, Stan Matwin, and Rita Orji

***Abstract*—The COVID-19 pandemic has affected people's lives in many ways. Social media data can reveal public perceptions and experience with respect to the pandemic, and also reveal factors that hamper or support efforts to curb global spread of the disease. In this paper, we analyzed COVID-19-related comments collected from six social media platforms using Natural Language Processing (NLP) techniques. We identified relevant opinionated keyphrases and their respective sentiment polarity (negative or positive) from over 1 million randomly selected comments, and then categorized them into broader themes using thematic analysis. Our results uncover 34 negative themes out of which 17 are economic, socio-political, educational, and political issues. 20 positive themes were also identified. We discuss the negative issues and suggest interventions to tackle them based on the positive themes and research evidence.**

## I. Introduction

Emerging infectious diseases are responsible for many deaths and disabilities globally [1]. Evidence shows that at least 43 million people contracted the H1N1 flu worldwide within 12 months of the pandemic which, in turn, resulted in over 200,000 deaths [2], [3]. In addition, 770,000 HIV/AIDS-related deaths were reported in 2018 alone, with over 37 million people infected globally [4]. The latest emerging infectious disease, COVID-19 [5], [6], has already infected over 15.7 million people worldwide, with a mortality of at least 640,000 as of July 25, 2020 [7]. Emerging infectious diseases have also been shown to inflict significant burden on economies and public health systems [8]–[10]. For example, global health systems are struggling to cope with the COVID-19 pandemic, while unemployment/job losses, reduced income/productivity, and business closures are prevalent among individuals and organizations due to the lockdown measures imposed by governments. To understand public perceptions toward the pandemic, social media data can provide the required insights from a global perspective [11].

Social media has been a major and rich data source for research in many domains, including health, due to its 3.8 billion active users [12] from diverse geographic locations across the globe. For instance, researchers analyzed user comments extracted from social media platforms (such as Facebook, Twitter, Instagram, discussion forums, etc.) to uncover insights about health-related issues (e.g., mental health [13], [14], substance use [15], [16], diseases [17]–[20], etc.), political issues (e.g., elections [21]–[24]), and business-related issues (e.g., customer engagement [25], [26]). With respect to COVID-19, social media comments can reveal public opinions about governments and health organizations' response to the pandemic, as well as economic, health, social, political, physical, and psychological impact of COVID-19 on global populations in line with the factors affecting efforts to limit the spread of the disease either negatively or positively.

In this paper, we apply natural language processing (NLP) to analyze COVID-19-related comments from six social media platforms (Twitter, Facebook, YouTube, and three online discussion forums) to uncover issues surrounding the pandemic based on public perceptions. NLP is a widely used method for extracting insights from unstructured texts, such as social media data and clinical texts (e.g., electronic health records [27] and patient journals [28]). Our research contributions are the following:

1) We implement context-based NLP for detecting relevant and opinionated themes or keyphrases from social media comments related to the COVID-19 pandemic.
2) We identify negative and positive themes that capture public opinions about the pandemic. Our results reveal 34 negative themes out of which 17 are economic, socio-political, educational, and political issues. 20 positive themes were also identified.
3) We suggest interventions to tackle the negative issues.

This work was supported by the NSERC Discovery Grant awarded to Dr. Rita Orji.

O. Oyebode, C. Ndulue, D. Mulchandani, B. Suruliraj, A. Adib, E. Milios, S. Matwin, and R. Orji are with the Faculty of Computer Science, Dalhousie University, Halifax NS, B3H 4R2, Canada (e-mail: oladapo.oyebode@dal.ca; cndulue@dal.ca; dn619055@dal.ca; banuchitra.suruliraj@dal.ca; ashfaq.adib@dal.ca; eem@cs.dal.ca; stan@cs.dal.ca; rita.orji@dal.ca).

F. A. Orji is with the Department of Computer Science, University of Saskatchewan, Saskatoon SK, S7N 5C9, Canada (e-mail: fidelia.orji@usask.ca). S. Matwin is also with the Institute of Computer Science, Polish Academy of Sciences, Warsaw, Poland.

The interventions, which are based on the positive themes and research evidence, will inform and help governments, organizations, and individuals to minimize the spread and impact of COVID-19, and to respond effectively to future pandemics.

## II. Related Work

Over the years, social media has been a rich source of data for health informatics research [29]. Natural language processing (NLP) techniques have been widely used for analyzing social media comments and clinical texts (such as identifying health-related and psychosocial issues with respect to the COVID-19 pandemic [30]).

The lexicon-based NLP technique was used to detect the prevalence of keywords indicating public interests in e-cigarette, marijuana, influenza, and Ebola using social media data; while Latent Dirichlet Allocation (LDA) technique was used to retrieve topics from the corpus [31]. LDA has also been utilized to extract latent topics from COVID-19-related comments posted on social media [32]. Also, the Natural Language Toolkit (NLTK) was used by [33] to identify top collocated n-grams (bigrams and trigrams) from clinical emails.

Furthermore, a custom topic modeling technique, called Ailment Topic Aspect Model, was employed to generate latent topics from Twitter data with the aim of identifying mentions of ailments of interest, including allergies, obesity, and insomnia [34]. The non-negative matrix factorization is another topic modeling technique used in health informatics research to extract topics from social media data [35]. A third-party tool for text mining, called KH-Coder, has also been used to explore potential topics related to H1N1-related advice, vaccine, and antiviral uptake in the United Kingdom based on Twitter data [36]. The machine learning-based NLP was utilized to analyze unstructured clinical notes to predict hospital readmissions for COPD patients [37] and perform sentiment analysis of user comments on mental health apps [38]. None of the techniques above considered contextual text analysis which can yield more meaningful and relevant outcomes.

To demonstrate the significance of contextual text analysis, Dave *et al.* conducted experiments to compare non-contextual N-Gram Chunking technique and the contextual Part-of-Speech (POS) chunking technique [39]. Rather than just extracting n-grams, the POS chunking method considers context of words by using regular grammars or POS patterns that specify how sentences should be deconstructed into keyphrases of interest. Their results show that systems using the POS chunking technique extracted relevant features (keyphrases) and outperformed systems adopting N-Gram chunking for feature extraction. We extend this approach with enhanced part-of-speech (POS) patterns tailored to our goal, chunking and CoNLL IOB tagging, as well as keyphrase transformation and sentiment scoring.

## III. Methodology

This paper aims to investigate public opinions regarding the COVID-19 pandemic and the impact of the disease on their lives. To achieve this objective, we utilize the following well-established computational techniques:

1) We developed programs or scripts to mine user comments related to COVID-19 from six social media platforms.
2) We preprocessed the data using NLP techniques.
3) We applied a seven-stage context-aware NLP approach to identify opinionated and meaningful keyphrases or themes from the comments.
4) We applied thematic analysis to iteratively categorize related keyphrases identified in step 3 above into broader themes or categories.

### A. Data Collection

A total of 47,410,795 COVID-19-related comments were collected across six social media platforms (Twitter, YouTube, Facebook, PushSquare.com, Archinect.com, and LiveScience.com), as described below:

1) **Twitter**: We built a console application to mine 47,249,973 tweets in real-time using the Twitter Streaming API [40] and C# programming language. The program targets tweets from the following hashtags: *#COVID19, #COVID, #ncov2019, #Covid_19, #StopTheSpread, #CoronaVirusUpdates, #StayAtHome, #selfquarantine, #COVID-19, #COVID—19, #panicbuying, #caronavirusoutbreak, #CoronaCrisis, #SocialDistancing, #cronovirus, #CoronaVirusUpdate, #MyPandemicSurvivalPlan, #Quarantined, #pandemic, #CoronavirusPandemic, #Coronavid19, #coronapocalypse, #QuarantineLife, #CoronaVirus, #QuarantineAndChill,* and *#CoronavirusOutbreak*.
2) **YouTube**: We wrote a Python script to automatically extract 111,722 user comments linked to 2,939 COVID-19-related videos using the YouTube Data API [41]. The keywords used for the video search include *covid-19*, *covid19*, and *coronavirus*.
3) **Facebook**: We adopted a semi-automatic technique to extract comments due to search restrictions imposed by Facebook. We first obtained 91 groups and 68 pages related to COVID-19 manually using the following keywords: *COVID*, *COVID-19*, and *Coronavirus*. Afterwards, we developed a Python script to retrieve 8,382 and 777 comments from the pages and groups respectively.
4) **Discussion forums**: We collected 18401, 20747, and 793 user comments from COVID-19-related threads on *PushSquare.com* [42], *Archinect.com* [43], [44], and *LiveScience.com* [45] respectively using Python scripts.

### B. Data Preprocessing

To clean the data and prepare it for keyphrase extraction, we apply the following preprocessing steps using NLP techniques implemented using Python:

1) Remove mentions, URLs, and hashtags
2) Expand contractions (such as replacing “couldn’t” with “could not”)

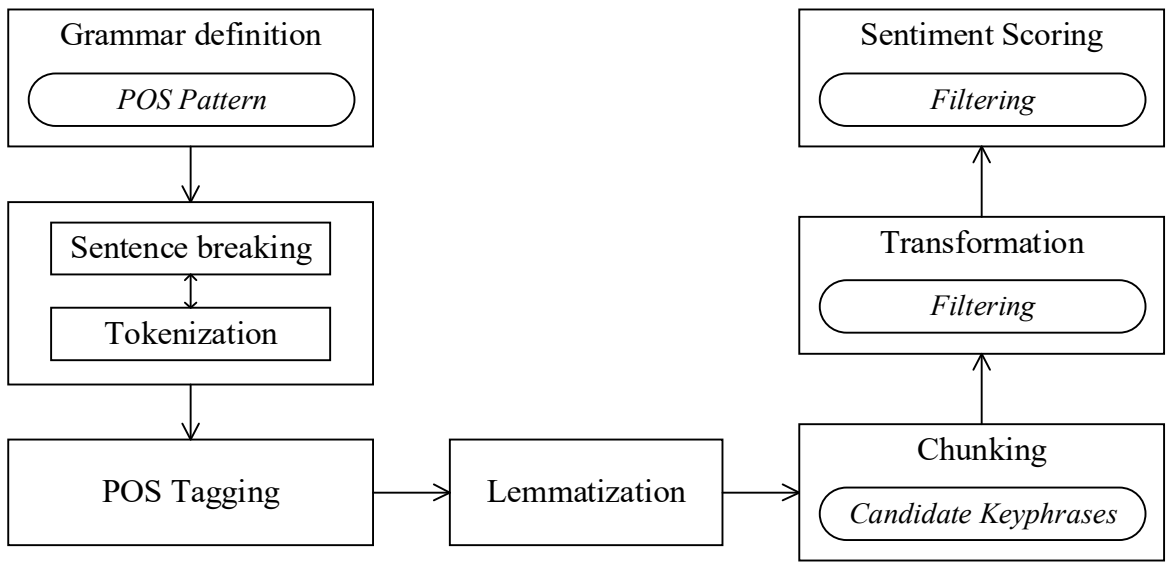

Fig. 1. A context-aware NLP approach for extracting keyphrases.

3) Replace HTML characters with Unicode equivalent (such as replacing “&” with “&”)
4) Remove HTML tags (such as “<div>”, “<p>”, etc.)
5) Remove special characters that are not required for sentence boundary detection
6) Compress words with repeated characters (such as compressing “poooool” to “pool”)
7) Convert slangs to English words using relevant online slang dictionaries [46], [47]
8) Remove words that are numbers

After applying the above steps on the data, and removing non-English comments (identified using the *langdetect* Python library [48]) and duplicate comments, the total number of comments reduced to 8,021,341. We randomly selected about 13% of these comments (n=1,051,616) to form the corpus used in this paper.

## C. Keyphrase Extraction

To extract meaningful and opinionated keyphrases/themes which are words or phrases representing topical content of each document (or comment) in our corpus, we utilized a context-aware NLP approach. This approach extends the version adopted by [39] with enhanced part-of-speech (POS) patterns tailored to our objective, chunking (in conjunction with CoNLL IOB tagging [49]), as well as transformation and sentiment scoring stages. In subsequent subsections, we describe the components of our architecture, as shown in Fig. 1. In Fig. 2, we present an algorithm (based on the architecture) which accepts a regular grammar and our corpus as parameters and returns keyphrases of interest as output. The

TABLE I
PART-OF-SPEECH (POS) TAGS AND DESCRIPTION

| POS Tag | Description | POS Tag | Description |
|---|---|---|---|
| NN | Noun (singular) | VB | Verb (base form) |
| NNS | Noun (plural) | VBD | Verb (past tense) |
| NNP | Proper Noun (singular) | VBG | Verb (gerund or present participle) |
| NNPS | Proper Noun (plural) | VBN | Verb (past participle) |
| JJ | Adjective | VBP | Verb (non-3rd person singular present) |
| JJR | Adjective (comparative) | VBZ | Verb (3rd person singular present) |
| JJS | Adjective (superlative) | DT | Determiner |
| IN | Preposition or subordinating conjunction | | |

```
KeyphraseExtractor(grammar, corpus) {
   for each document in corpus {
      sentences ← extract_sentences(document)
      pos_tagged_tokens_all ← empty
      for each sentence in sentences_list {
         tokens ← extract_tokens(sentence)
         tagged_tokens ← assign_pos_tag(tokens)
         lemmatized_tokens_tagged ← lemmatize(tagged_tokens)
         append lemmatized_tokens_tagged to pos_tagged_tokens_all
      }
      chunker ← create_syntactic_parser(grammar)
      chunks ← chunker.parse(pos_tagged_tokens_all)
      candidate_keyphrases ← generate_keyphrases(chunks)

      for each keyphrase in candidate_keyphrases {
         if keyphrase is a stopword then
            remove keyphrase from candidate_keyphrases
         else {
            keyphrase_new ← strip_selected_stopwords(keyphrase)
            if keyphrase_new is empty then
               remove keyphrase from candidate_keyphrases
            else if length(keyphrase_new) > 10 then
               remove keyphrase from candidate_keyphrases
            else {
               sentiment_score ← get_sentiment_score(keyphrase_new)
               sentiment_polarity ← get_polarity(sentiment_score)
               if sentiment_polarity = "neutral" then
                  remove keyphrase from candidate_keyphrases
               else
                  set keyphrase to keyphrase_new in candidate_keyphrases
            }
         }
      }
   }
   return candidate_keyphrases
}
```

Fig. 2. The *KeyphraseExtractor* algorithm based on the context-aware NLP approach.

algorithm was implemented using Python.

### 1) Grammar definition

We defined a regular grammar (see below) which is a set of rules composed of POS patterns that describe how the syntactic units of each document in our corpus are deconstructed into their constituents or parts. The grammar captures the context of each comment and the opinions/sentiments expressed using nouns, adjectives, and verbs. Research revealed that nouns are crucial for detecting the context of a conversation [50], while both adjectives and verbs are significant for sentiment classification [51].

*Grammar:* { <DT>? <JJ.*>* <NN.*>* <VB.*>? (<IN>? <DT>? <JJ.*>* <NN.*>*)? }

The regular grammar above is composed of patterns of POS tags from the well-established Penn Treebank Tagset [52], [53]. For instance, the <NN.*> pattern matches any type of noun (see Table I), <JJ.*> matches any type of adjective, <VB.*> matches any type of verb, <IN> matches a preposition or subordinating conjunction, and <DT> matches a determiner. We considered determiners and prepositions since they usually occur together with nouns and adjectives in sentences (e.g., *public concern* ***about the*** *virus*). Also, the “*” symbol after a POS pattern refers to “zero or more occurrences”, while “?” refers to “zero or one occurrence”.

### 2) Sentence breaking and Tokenization

Next, each document is separated into unique sentences. To achieve this, we utilized a robust unsupervised algorithm (within the Python NLTK’s *tokenize* library [54]) which considers collocations, punctuations, capitalizations, and abbreviations in determining sentence boundaries within each document. Afterwards, each sentence is further broken down into words or tokens in preparation for POS tagging.

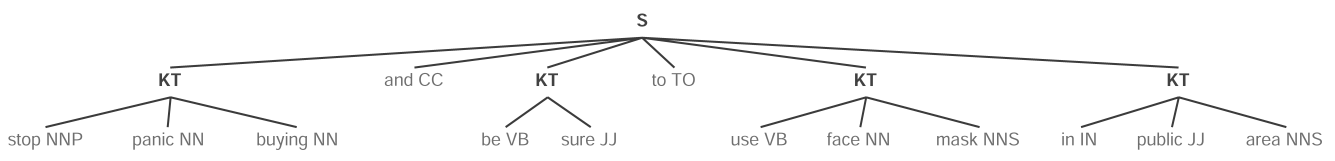

Fig. 3. A sample parse tree illustrating the output of the chunker.

#### 3) POS Tagging

Each token is assigned a POS tag (within the Penn Treebank Tagset) denoting its part of speech in the English language. For example, tokens in the following sentence *"Stop panic buying and be sure to use face masks in public areas"* are tagged as follows: *[('Stop', 'NNP'), ('panic', 'NN'), ('buying', 'NN'), ('and', 'CC'), ('be', 'VB'), ('sure', 'JJ'), ('to', 'TO'), ('use', 'VB'), ('face', 'NN'), ('masks', 'NNS'), ('in', 'IN'), ('public', 'JJ'), ('areas', 'NNS')]*.

#### 4) Lemmatization

Next, each tagged token is lemmatized or converted into its root word based on its part of speech. Prior to lemmatization, we converted the tokens or words to lowercase. Lemmatization is achieved by using the English vocabulary and conducting morphological analysis of words [55]. Hence, a root word is the dictionary form of the original word. By converting the tokens to their root form, we harmonized similar words while preserving their meaning. For instance, the following verb words "seen" and "sees" are converted to their root form – "see". Referring to our previous sample tagged tokens, the output of the lemmatization stage is: *[('stop', 'NNP'), ('panic', 'NN'), ('buying', 'NN'), ('and', 'CC'), ('be', 'VB'), ('sure', 'JJ'), ('to', 'TO'), ('use', 'VB'), ('face', 'NN'), ('**mask**', 'NNS'), ('in', 'IN'), ('public', 'JJ'), ('**area**', 'NNS')]*.

#### 5) Chunking

Next, we created a chunker that uses the regular grammar defined above to match phrases comprising an optional determiner, followed by zero or more of any type of adjective, zero or more of any type of noun, zero or one of any type of verb, and an optional component. This component consists of an optional preposition, followed by an optional determiner, zero or more of any type of adjective, and zero or more of any type of noun. Using our previous example, the chunker produces the parse tree in Fig. 3, showing the key terms (KT) that match the grammar specified.

To generate the candidate keyphrases, we first converted the parse tree (or chunks) generated by the chunker for each document into a CoNLL IOB format. An IOB (Inside-Outside-Beginning) tag specifies how a key term functions in the context of a phrase – whether the term *begins* (B-KT), is *inside* (I-KT), or *outside* (O-KT or O) the phrase [49]. Next, we iteratively group terms that are part of a keyphrase (i.e., B-KT and I-KT) and stops when a term that does not belong to the keyphrase (i.e., O-KT or O) is encountered.

For example, the CoNLL IOB format of the parse tree in Fig. 3 gives *[('stop', 'NNP', 'B-KT'), ('panic', 'NN', 'I-KT'), ('buying', 'NN', 'I-KT'), ('and', 'CC', 'O'), ('be', 'VB', 'B-KT'), ('sure', 'JJ', 'I-KT'), ('to', 'TO', 'O'), ('use', 'VB', 'B-KT'), ('face', 'NN', 'I-KT'), ('mask', 'NNS', 'I-KT'), ('in', 'IN', 'B-KT'), ('public', 'JJ', 'I-KT'), ('area', 'NNS', 'I-KT')]*. By iteratively grouping the B-KT and I-KT terms, the following keyphrases emerged: *"stop panic buying"*, *"be sure"*, and *"use face mask in public area"*.

TABLE II
CRITERIA FOR SENTIMENT CLASSIFICATION

| Condition | Sentiment Polarity |
|---|---|
| $S_s > 0.05$ | Positive |
| $S_s < -0.05$ | Negative |
| $S_s \geq -0.05$ and $S_s \leq 0.05$ | Neutral |

#### 6) Transformation and Filtering

In this stage, we removed keyphrases that are stopwords (i.e., common words, such as *about*, *the*, *from*, *there*, *had*, *can*, etc.) from our list of candidate keyphrases. We also removed selected stopwords from the start, end, and within keyphrases while preserving their meaning. For example, *"be sure"* will be filtered out since *"be"* and *"sure"* are included in our pre-defined list of stopwords that cannot start nor end a keyphrase. Third, we removed keyphrases whose length exceed ten. While previous research retained only keyphrases up to length six [39], we extended our threshold to ten to avoid losing important keyphrases that would enrich insights from this research.

#### 7) Sentiment scoring and Filtering

In line with our objective to keep only opinionated candidate keyphrases (i.e., keyphrases with "negative" or "positive" sentiment polarity [56]), we assigned a sentiment score $S_s$ ranging from –1 to +1 to each keyphrase using the popular VADER lexicon-based algorithm [57]. Afterwards, we filtered out non-opinionated or "neutral" keyphrases using the criteria summarized in Table II. For example, the $S_s$ for *"stop panic buying"* and *"use face mask in public area"* are –0.6705 and 0.1027 respectively; hence, will be retained since they are opinionated. The neutral score ranges from –0.05 and +0.05 based on the outcome of the experiments conducted by [57].

### D. Categorizing Keyphrases

Next, the final keyphrases or themes were manually categorized into broader themes (an approach also used by [33] to categorize phrases) by four reviewers. The reviewers were divided into two teams – T1 and T2. T1 consists of two reviewers who were tasked with grouping the negative keyphrases, while T2 comprises the two other reviewers who grouped the positive keyphrases. Each reviewer iteratively examines the keyphrases and continues to group them until no new category emerges (i.e., saturation level is reached). The percentage of agreement between reviewers in T1 is 98%, while that of T2 is 99.3%.

## IV. RESULTS

### A. Negative Keyphrases

A total of 427,875 unique negative keyphrases were generated from our large corpus. Fig. 4 shows some of the keyphrases in our results and their dominance (the gray bubbles of varying sizes illustrate dominance based on frequency of occurrence). In decreasing order, *death* is the most dominant keyphrase (n=10,187), followed by *die* (n=7,240), *fight* (n=5,891), *bad* (n=3,808), *kill* (n=3,668), *lose* (n=3,631), *pay* (n=3,486), *leave* (n=3,234), *crisis* (n=2,783), *hard* (n=2,720), *worry* (n=2,476), *sick* (n=2,314), *sad* (n=2,129), etc. Some non-unigram keyphrases in our results include *self isolation*, *difficult time*, *life at risk*, *death toll rise*, *conspiracy theory*, *become infected*, *spread misinformation*, *panic buy*, *lack of leadership*, *no social distancing*, *travel restriction*, *spread fake news*, *in time of uncertainty*, *public health emergency*, *biological weapon*, *desperate time call for desperate measure*, *contagious disease*, *hospital overwhelm*, *take advantage of crisis*, *suffer from underlie medical condition*, etc.

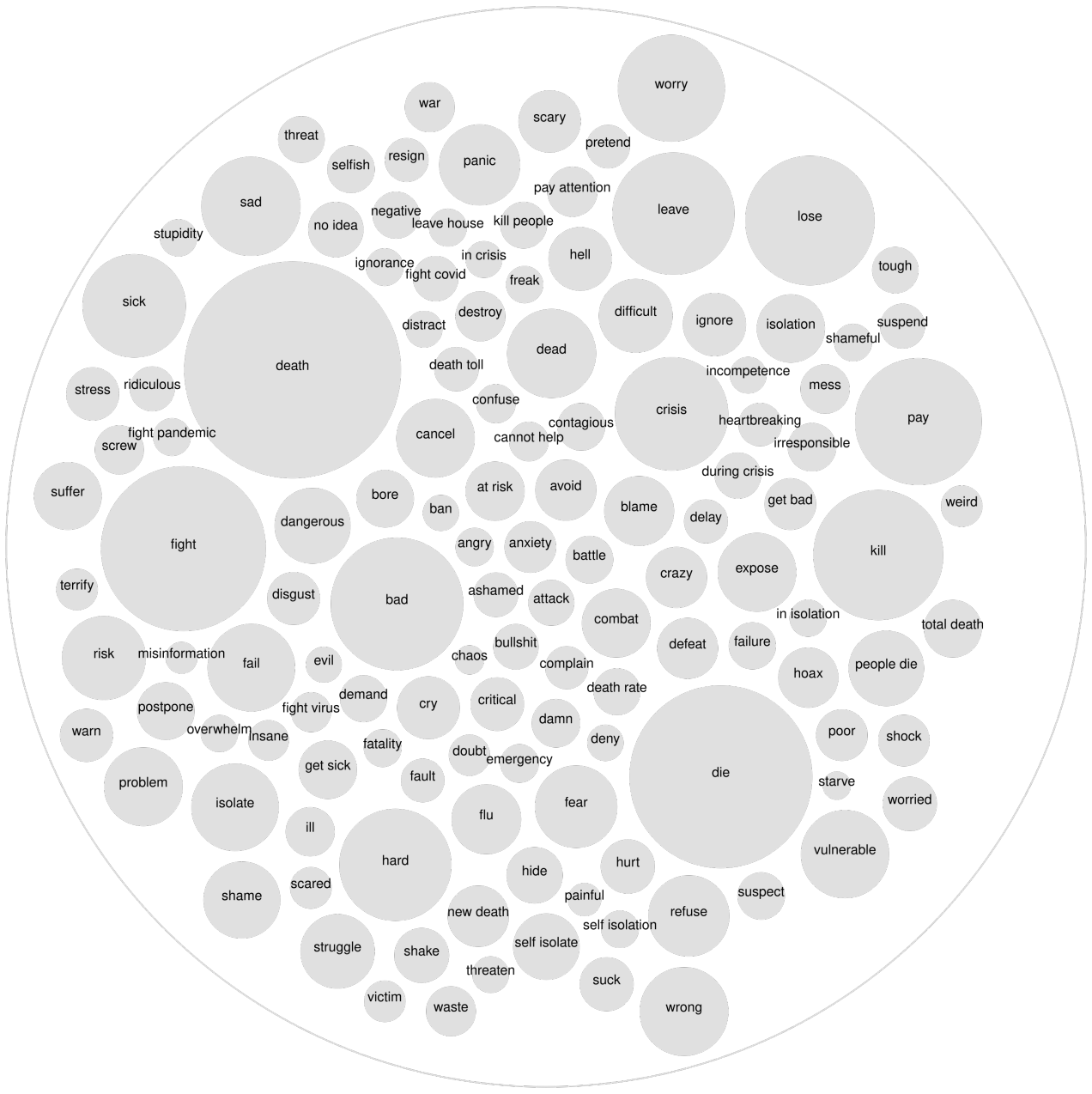

Fig. 4. Sample negative keyphrases and their frequency of occurrence (a larger bubble size illustrates higher dominance).

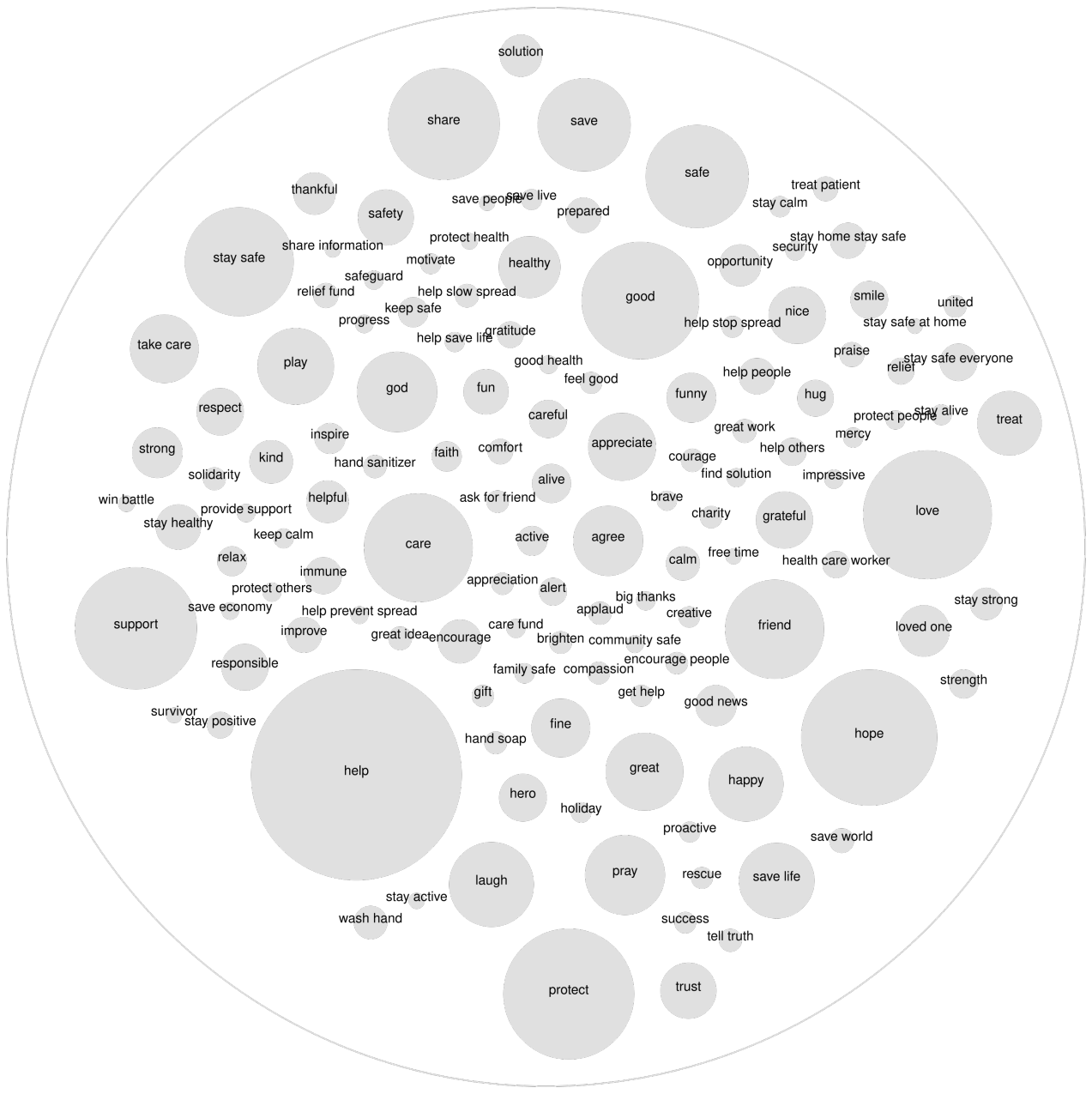

Fig. 5. Sample positive keyphrases and their frequency of occurrence (a larger bubble size illustrates higher dominance).

### *B. Positive Keyphrases*

On the other hand, 520,685 unique positive keyphrases emerged from our corpus. Some of the keyphrases and their dominance based on frequency are illustrated in Fig. 5. The most dominant keyphrase, in decreasing order, is *help* (n=18,498), followed by *hope* (n=7,708), *protect* (n=7,130), *love* (n=6,895), *support* (n=6,198), *good* (n=5,740), *share* (n=5,187), *care* (n=4,917), and *stay safe* (n=4,917). Other keyphrases in our results include *stay healthy*, *gratitude*, *relief fund*, *help slow spread*, *solidarity*, *ask for friend*, *encourage people*, *stay calm*, *great initiative*, *fresh air*, *use hand sanitizer*, *artificial intelligence*, *support business*, *keep safe distance*, *practice good hygiene*, *pray at home*, *play video game*, *use defense production act*, *protect public health*, *encourage social distancing*, *free webinar*, etc.

### *C. Keyphrase Categories*

After grouping related keyphrases into broader themes or categories (as discussed in the Methodology section), 34 negative and 20 positive categories emerged. We refer to these categories as "themes", and the keyphrases under each category as "subthemes" in the remaining parts of this paper. The 34 negative themes are further distributed into health-related issues, economic issues, psychosocial issues, socio-political issues, social issues, educational issues, and political issues. In this paper, we focused on 17 negative themes mapped to economic, socio-political, educational, and political issues (see Table III). As shown in Fig. 6, the top 5 negative themes based on number of user comments is *Concerns about social distancing and isolation policies* (n=8,872), followed by *Misinformation* (n=2,223), *Political influence* (n=1,640), *Financial issues* (n=1,622), and *Poor governance* (n=1,559). Fig. 6 also shows the number of subthemes under each theme.

Furthermore, Fig. 7 shows the 20 positive themes and the corresponding number of subthemes and comments. Based on number of comments, *Public awareness* (n=22,749) emerged as the top theme, followed by *Spiritual support* (n=12,130), and *Encouragement* (n=5,244). Other themes include *Charity* (n=942), *Entertainment* (n=798), *Gratitude* (n=758), *Development of curative solutions or treatments* (n=653), *Advocacy for increased testing* (n=341), *Physical activity* (n=285), *Cleaner environment* (n=278), etc.

TABLE III
THEMES, DESCRIPTION, AND SAMPLE COMMENTS

| Theme | Description | Sample comments |
|---|---|---|
| ECONOMIC ISSUES | | |
| Job and Business-related crisis | Loss of jobs, shortage of open jobs, reduction in salary or wage, reduction in organization's revenue, etc. | *"…tens of thousands of hotel staff being temporarily* ***laid off*** *in Marriott's US operations alone -- and several hotel CEOs and executive staffs are* ***cutting their salaries by 50 or forgoing it altogether****. So sad to see."*[a] *[C993]* |
| Challenging living condition | Reduced standard of living or difficulty in satisfying essential needs. | *"COVID19 has created* ***new hardships*** *for the 14.3 million U.S. households already experiencing* ***food insecurity****…" [C21]* |
| Economic downturn | Struggling global economies, as revealed through stock market crash, reduction in currency value, reduced GDP, etc. | *"The **Stock Market is falling**! Emergency! ...both **investors AND non investors are affected by a stock market crash**." [C12900]* |
| Financial issues | Difficulty in meeting financial obligations such as paying bills (e.g., rent, mortgage, credit card, phone, etc.), paying workers' salaries/wages, etc. | *"Starting a new week in despair... Little small business I administrator for will not be able to operate in a lockdown. **No money, no rent**. We could be homeless in weeks. Very scary uncertain times..." [C180]* |
| Shortage of essential products or items | Inadequate supply of essential items such as personal protective equipment (e.g., face mask and protective gear), testing kits, ventilators, food, hand sanitizers, toilet paper, and blood (in blood banks) | *"The **national shortage of personal protective equipment PPE** is making pandemic worse. Our doctors, nurses first responders cannot effectively save lives if they cannot protect their own health get sick..." [C4468]* |
| Flight cancellations | Concerns over unexpected cancellation of domestic and international flights, including refund issues | *"Shame on you for **cancelling all flights leaving people stranded** whilst **refusing to refund your customers**. You have added to the significant hardship and stress already caused by covid19. Absolutely disgraceful behaviour. You should hang your heads in shame!" [C133]* |
| | SOCIO-POLITICAL ISSUES | |
| Concerns about social distancing and isolation policies | Public concerns/reactions toward lockdown, social distancing, or isolation policies imposed by governments, such as viewing them as difficult, harsh, worthless, not well-planned, etc. | *"Inaccurate Virus Models from COVIDActNow are panicking officials into **ill-advised lockdowns**. There was **no reason to go into lockdown** and destroy the economy. This could have been managed with accurate information and NPIs." [C940]* |
| Controversy over precautionary measures | Controversy over certain precautionary or safety guidelines, such as wearing of face masks, gloves, etc. | *"You will **look very stupid wearing a face mask and gloves** in a party, restaurant or wherever..." [C463]*<br><br>*"Okay. I think **wearing a mask is kind of over exaggerating things**." [C20]* |
| Lack of preparedness | Associate disease spread or pandemic to lack of preparedness by governments and health agencies | *"**Pathetic planning, total lack of preparation for COVID19**. More ventilators could have been in place weeks ago." [C788]* |
| Protests | Public protests during pandemic against government actions or due to other reasons | *"People **protest regime of Iran for covering up the truth on outbreak** and the way they bury bodies." [C4329]* |
| Risk of spread at detention centres | Concerns over spread of disease in prisons or detention centres due to the welfare issues and overcrowding | *"Already **27 people in the UK across 14 prisons have tested positive for COVID19 and 2 have died**. Conditions in **prisons are unsanitary and unhygienic**. We need an immediate release of all people in prison to prevent further death." [C2411]* |
| | EDUCATIONAL ISSUES | |
| Disruption in Education | Negative impact or effects of the pandemic on education, including closure of schools, suspension of daycare for kids, etc. | *"Today, the Baker-Polito Administration announced a three-week **suspension of school operations for educational purposes at all public and private elementary and secondary K-12 schools** in the Commonwealth beginning Tuesday, March 17." [C5568]* |
| Misinformation | Concerns over false or unfounded information regarding the disease or pandemic | *"Hey, this is **false information** about Covid19. Do you not have a sense responsibility towards us during this time? Remove this tweet, it is **spreading fake information regarding the pandemic**." [C689]* |
| Knowledge gap | Insufficient knowledge on how to deal with the disease or curb spread | *"Staggered by the public's **lack of knowledge**... Boris did not shut the whole hospitality industry down so you could congregate in the supermarket. Please shop on your own and leave the family at home. You are putting lives at risk." [C270]* |
| | POLITICAL ISSUES | |
| Poor governance | Blaming the governments of various countries due to their incompetence, poor health infrastructure, etc. | *"...but given his **complete incompetence** in handling pandemic...it is worst **failure of leadership** in human history. **Many Americans will die because of his incompetence**" [C398]* |
| Political influence | Complaints about playing politics with pandemic (e.g., politicians raising conspiracy theories or propaganda to score political points, covering up pandemic impact, etc.) | *"Blah blah blah! It started the same with Italy and now they are crying of massive deaths caused by Leaders in Italy now regretting of **imposing politics on it** instead of taking required measures. I can see we are following the same path." [C733]* |
| Infrastructural issues | Technical or infrastructural issues during pandemic, especially lack of internet access or poor internet quality, unstable mobile reception, etc. | *"**No internet**, almost **no cellular reception**. It's no way to spend the vegas shutdown" [C3224]* |

[a] Comments are included verbatim throughout the paper, including spelling and grammatical mistakes.

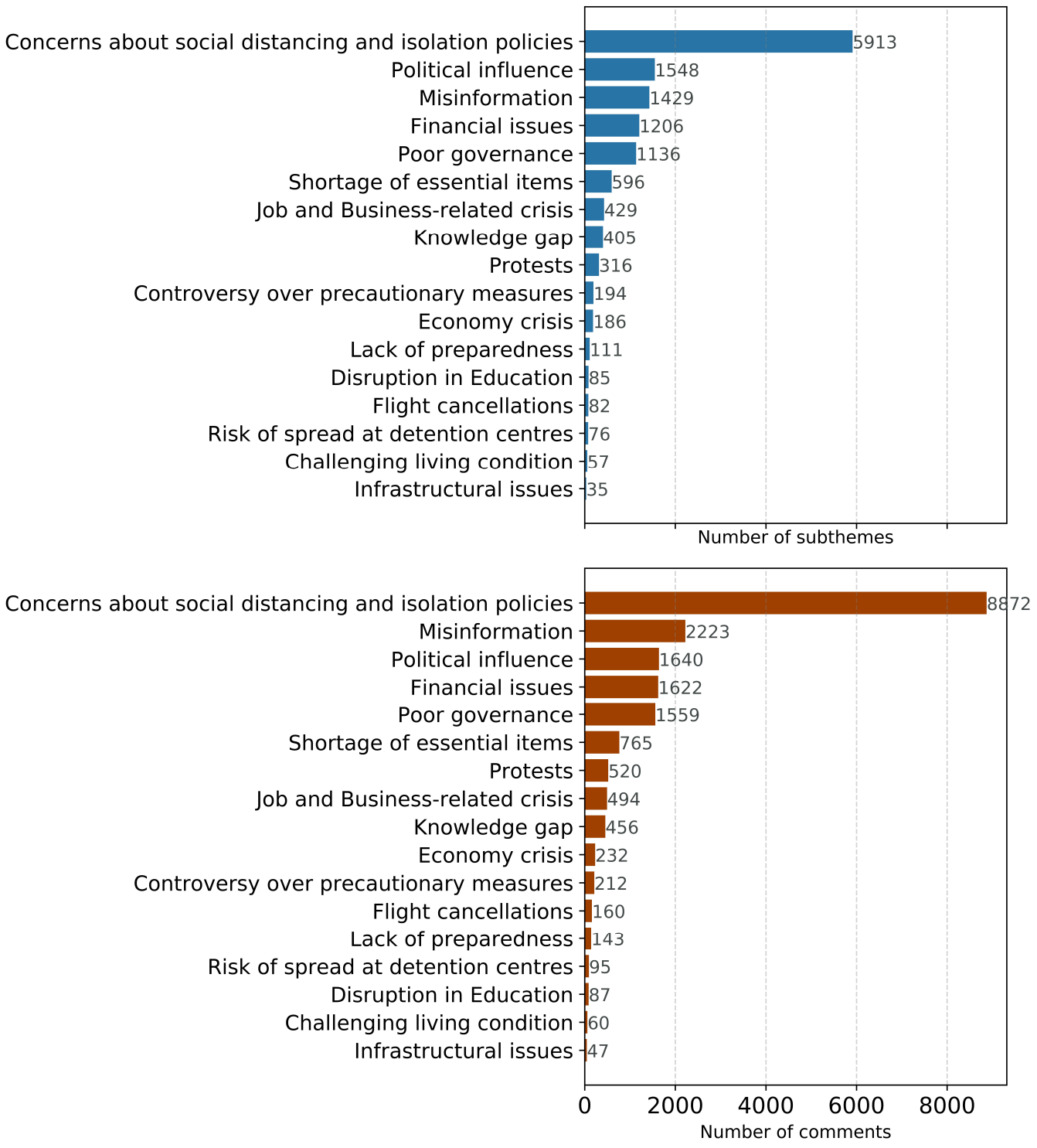

Fig. 6. The top chart shows negative themes and the corresponding number of subthemes, while the bottom chart shows the total number of user comments for each theme.

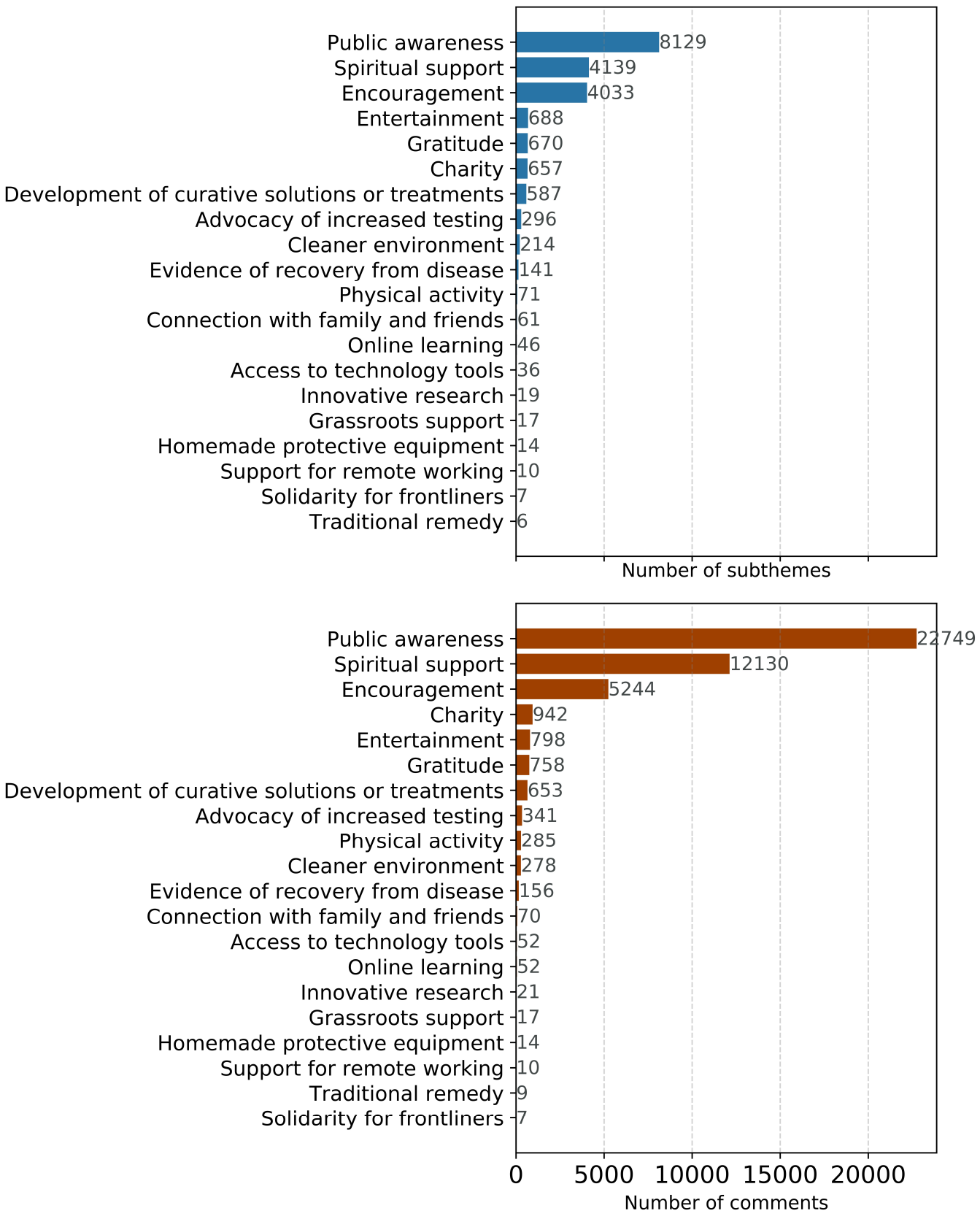

Fig. 7. The top chart shows positive themes and the corresponding number of subthemes, while the bottom chart shows the total number of user comments for each theme.

## V. Discussion

Our results revealed various negative and positive themes representing public opinions about the pandemic, as well as impact of COVID-19 on people and institutions in line with the factors affecting efforts to limit the spread of the disease either negatively or positively. In this section, we discuss the negative issues (see Table III) and suggest interventions based on the positive themes (see Fig. 7) and research evidence.

### A. *Negative issues regarding COVID-19 pandemic*

#### 1) *Economic issues*

Based on our findings, the COVID-19 pandemic led to unemployment, low revenue or losses for business, low supply of essential items, challenging living condition, economic downturn, and financial crisis.

##### a) *Job- and Business-related crisis*

In line with our findings (see sample comments below), research shows that the pandemic triggered massive global unemployment crisis [58]–[60] where people are losing jobs or unable to get one. This is due to lockdowns and reduced consumer spending which led to businesses/companies experiencing low income/revenue and losses as many near bankruptcy, shutdown temporarily, or likely to go out of business [61]–[64].

> *"...**job layoffs are soaring faster** than any time in recorded history...This looks bad and it is bad. The **worst jobless claims in U.S. history** means the economy has fallen into the abyss." [C9100]*

> *"My **job is shutdown**; my husband **job is shutdown**...How am I supposed to pull this off? There is **NO income**. We have 4 children including an 8-week-old baby. I need help NOW." [C7119]*

##### b) *Economic downturn*

Based on our findings, the pandemic pushes global economies toward recession as stock market indices crashes, as shown in the sample comment below. Evidence shows that the COVID-19 pandemic negatively impacted stock markets more forcefully than any other disease outbreak in history [65]. For example, primary sectors (e.g., agriculture and petroleum and oil), secondary sectors (e.g., manufacturing), and other sectors (e.g., finance, food, real estate, tourism, and transportation sectors) driving stock market indices experienced various challenges (such as supply chain disruption, revenue crash, transaction halt, etc.) compounded by lockdown and social isolation policies aimed to curb COVID-19 spread [66].

> *"Our 250 economists have updated our global forecasts. **Coronavirus will inflict a short, sharp global recession**. We expect **2020 world growth to drop to zero**. In Q1, we see the **global economy shrinking faster than in the financial crisis**" [C10002]*

##### c) *Shortage of essential items*

People lamented shortage of food items, toiletries including hand sanitizers, and personal protective equipment (e.g., face masks, protective gear and garment, etc.) necessary to prevent contracting the disease. In addition, public health centres and hospitals experience shortage of testing kits and ventilators which hampered efforts to identify COVID-19 cases and keep patients alive. Also, blood shortages were reported in blood banks and lockdown measures may prevent many people from donating blood. Our findings (see sample comments below) align with research which confirms critical supply shortages of the items highlighted above [67]–[71].

> *"U.S. cities have **acute shortages of masks, test kits, ventilators** as they face coronavirus threat" [C11119]*

> *"**Acute shortage of blood in the blood banks**...Blood donations needed during & after coronavirus pandemic" [C7999]*

> *"Is anybody else having a **food shortage in their grocery stores**? My hometown stores are about completely empty." [C4442]*

*d) Challenging living condition and Financial Issues*

As shown in the comments below, people experienced difficulty providing for their families or meeting their needs such as paying bills (e.g., rent, mortgage installment, credit card payment, phone bill, etc.) and buying sufficient food, as a result of job losses and the strict lockdowns which impose financial hardship on people, including owners of small businesses (such as restaurants and cosmetics shops) and hustlers. Many organizations are unable to pay employees' full salary due to financial constraints caused by the pandemic and resort to half salaries or job cuts (see *[C621]*). Research has shown that households experienced food insecurity as a result of poor financial status caused by the COVID-19 pandemic [72].

> *"Please help us. Lost my job due to coronavirus shutting down my workplace. I have **no income**...**no rent**." [C9991]*

> *"...there are a **heartbreaking number of hungry Americans** posting their Venmo's and asking for help in this thread..." [C12883]*

> *"Today my company instituted across-the-board **pay cuts of 10-30%**, and **canceled merit increases**, **bonuses**, and **401(k) matches**..." [C621]*

*e) Flight cancellations*

Based on our findings, people lamented sudden cancellation of flights and difficulty in getting refunds from affected Airlines, as shown in the sample comment below. These cancellations are due to border closures and travel bans imposed by governments of many countries to curtail the importation of COVID-19; however, such actions inflicted much pain and distress to stranded passengers, as well as financial losses to the Airlines [73].

> *"Very disappointed how Etihad is handling COVID19. Not only did they **cancel all flights**, but it is legally impossible for me to travel given the **travel bans**. **Instead of refunding my money**, I am getting **credit that has restrictions to re-book by Sept.** How is this fair?" [C144]*

### 2) Socio-political issues

*a) Concerns about social distancing and isolation policies*

Our findings revealed public concerns over lockdown, social distancing, and isolation policies irrespective of their perceived benefits. Some of the concerns include: (i) people disrespectfully snubbing those who are not 6-feet away from them; (ii) social distancing/lockdown without financial support; (iii) implementing isolation/quarantine policies that contradict the World Health Organization's advice; (iv) human rights violation; (v) weak enforcement of lockdown policy; (vi) reliance on self-isolation without aggressive testing; (vii) ineffectiveness of isolation/lockdown in slums; (viii) devastating effect of social isolation on domestic violence victims; (ix) millions stranded due to lockdown and struggling to get food and water; (x) spike in anxiety and depression cases after lockdown announcement. Sample comments are shown below:

> *"...but of societal norms; so much so that there are now actual social distancing snobs who **look down on you if you're less than 6 feet away**. Will coronavirus kill all our humanity too?" [C7116]*

> *"This is exactly what I am afraid of since lockdown...**exploiting the crisis to strip us of our rights**." [C909]*

*b) Controversy over precautionary measures*

Precautionary measures, such as wearing face masks and gloves, generated controversies based on our findings. For example, some people think N95 masks with a valve can aid the spread of the virus from infected patients (see sample comment below), while some are concerned about the stigma attached to wearing masks.

> *"They tell the infected to wear a N95 mask which is 95 effective with no oils...half the masks have a **one way valve for the exhale which is unfiltered**. They are **trying to kill everyone**." [C193]*

*c) Lack of preparedness and Protests*

Furthermore, people highlighted lack of preparedness on the part of governments and health systems as a factor aiding the spread of the disease. Evidence shows that many countries and health authorities failed to rapidly perceive the threat posed by COVID-19 [74], [75], thereby allowing it to degenerate into a pandemic level that imposes hardship on world population. Therefore, it is unsurprising that there are protests in several countries, such as health workers and some essential workers protesting about shortage of protective equipment, citizens of developing countries protesting about lack of food and electricity during the lockdown, essential workers requesting for hazard pay during the pandemic, citizens protesting against their government's inactions toward protecting them from COVID-19, and so on. Below are sample comments:

> *"Another **nurses protest calling attention to the shortage of protective equipment**, and **rationing policies by hospitals**." [C299]*

> *"This is what is happening in Chile. **People protesting** because President…is **not taking the correct responsibility on Covid_19**. We need national quarantine!!" [C10384]*

> *"We are hungry, no food no light - you cannot tell us to stay indoors. Nigerians in parts of Lagos already **threatening to defy government lockdown directives with a protest in two days**" [C8886]*

*d) Risk of spread at detention centres*

Moreover, people raised concerns about risk of spreading COVID-19 disease in prisons and the need for decongestion, as shown in the sample comment below. Evidence already shows that incarcerated populations are vulnerable to infectious diseases, including COVID-19, due to unavoidable close contact (since prisons or detention centres are often overcrowded and poorly ventilated/sanitized) and poor healthcare access [76], [77].

> *"As cases increase in Texas, so do* ***concerns about the well-being of people in Texas prisons and jails.****" [C1555]*

### 3) Educational issues

#### a) Disruption in Education

Our findings revealed the disruptive effect of the COVID-19 pandemic on education globally, such as school closures. People are concerned about their children or wards' education including the cost implication of virtual learning put in place by schools, as well as children in rural areas who would be deprived of learning. This aligns with evidence highlighting the effect of school closures on 80% of children worldwide, and the worsened inequalities in educational outcomes between children in lower- and higher-income countries [78].

> *"There is still* ***NO date for schools to reopen*** *in the Capital." [C4440]*

> *"You are putting the most disadvantaged students at a further educational disadvantage..." [C91]*

> *"What kind of education are we getting? We haven't paid such* ***high fees for ZOOM kind of education****. Our* ***online education system is so sick and badly affecting our grades*** *and that's totally unfair!! 90k for this kind of education is way too worthy!" [C181]*

#### b) Knowledge gap

Furthermore, knowledge gap (in form of ignorance and lack of intelligence) on the part of leadership and society is another factor hampering the containment of COVID-19, based on our findings. As shown in the comments below, authorities are short of knowledge as regards what should be done, while people are ill-informed due to limited access to accurate and coherent information about the disease and preventive/control measures.

> *"Nursing, a primarily female profession, is under attack. The* ***CDC in ignorance says wear a bandana****.* ***THIS IS NOT PROTECTION****!* ***Lives at stake****! Hospitals have their heads in the sand. Please! Can you hear us?" [C818]*

> *"Some neighbors even want us out because they* ***think they would breathe this virus in the air and we're inside our own hostel****! This is* ***dangerous ignorance****!" [C6000]*

#### c) Misinformation

The proliferation of misinformation is impeding access to accurate information about COVID-19 that could have helped curb the spread of the disease and save lives. Misinformation which refers to false information or information with limited or without scientific evidence is one of the top 5 issues that emerged in our findings, and also reported by previous research [79]–[84]. Sample comment below reveals public concerns about misinformation regarding COVID-19:

> *"The* ***amount of fake news on all the health concerns regarding COVID19 is shocking****. Only person I trust with info is my cousin who is a doctor. She has just told me it DOESN'T last in the air, as long as you are two metres away and sneeze into a tissue you are fine!!" [C12010]*

### 4) Political issues

Elected governments or political appointees are central to decision making or governance that should improve the standard of living of people and assure their health and safety. However, people are concerned about widespread interference in COVID-19-related matters for political gain, based on our findings. In addition, they are concerned about the absence of strong leadership in the wake of the pandemic, and the poor state (or lack) of key public infrastructure (e.g., electricity, water, internet, healthcare facilities/centres, etc.). Research has shown that political beliefs and partisanship pose a significant limitation on the effectiveness of preventive measures (such as social distancing) [85]–[87]. Sample comments below reveal public opinion regarding these issues:

> *"Quite possibly the* ***worst governor in the country****. He's* ***hurting not only his own citizens but all Americans****, as all the* ***people on spring break and theme parks go home with covid19****." [C5777]*

> *"When authorities and armed forces asked you to self-quarantine with* ***no internet*** *and* ***no electricity (16 hours load shedding)*** *in Hunza ...?!" [C108]*

## B. Interventions for addressing the negative issues

To cushion the effect of economic issues on people, *"charity"* and *"grassroots support"* are important factors as revealed in our findings. Mobile technology can play a significant role in ensuring effective distribution of relief items. For example, GPS-enabled and multilingual mobile apps can help people to easily find food banks nearest to them. Moreover, government-funded or non-governmental charity organizations responsible for distributing economic relief to people can easily expand their reach or coverage and make delivery decisions based on data collected through these apps. For instance, people can indicate their needs through these apps and other information, such as location, age group, health condition, and whether they are in self-isolation (due to exposure to COVID-19). These apps can also be used to onboard volunteers who want to offer financial and material assistance and connect them to those in need. Furthermore, the data collected through these apps can further be analyzed using artificial intelligence (AI) techniques (such as machine learning or deep learning) to predict the communities that are in dire need of assistance. Besides technology usage, governments can budget for additional measures to protect the finances of people and businesses such as keeping people employed through financial partnership with employers, providing stimulus packages, and facilitating quick employment for the jobless [88]. Evidence shows that governments of some countries are adopting these measures to varying degrees [89].

Regarding shortage of items to protect people from the pandemic (such as face mask and hand sanitizers), a *"homemade protective equipment"* approach can be employed as a viable alternative, as revealed in our findings. Evidence shows that homemade masks, for example, can offer protection from COVID-19 transmission, in the event where medical masks are not available [90]. To address supply chain issues with respect to high demand and essential products, research suggests recovery strategies (such as increase in production shifts, use of spare capacity, emergency sourcing, bolstering capacity locally, and collaboration with supply chain partners) [91], [92].

Regarding concerns about social distancing and isolation policies imposed by governments, as well as controversies over precautionary measures suggested by health professionals, *"public awareness"* is a major and useful tool to address these issues, including misinformation, as revealed in our findings. Providing timely and accurate COVID-19-related information to people, and also connecting them to evidence-based resources and health professionals to resolve their questions or confusions, can be lifesaving. To reach a wider audience on a personalized basis, mobile- and voice-enabled chatbots equipped with real-time access to evidence-based and validated resources (such as approved safety measures by World Health Organization, as well as government-approved policies or guidelines) can be developed such that people can interact (in their own language) with the chatbots using their smartphones anytime. Difficult questions can be automatically channeled to health experts for responses within the same chat window. For those with traditional cellular phones, governments and local health agencies can partner with telecom firms to deliver COVID-19-related information directly to people's phones as a short messaging service (SMS) at regular intervals. In addition, official COVID-19-related channels on social media (such as [93]) supervised by health experts and local/international health organizations can provide accurate and frequent updates.

Regarding educational disruptions due to COVID-19, evidence shows that digital technologies are pedagogical tools that can enhance diverse forms of learning both within and outside the school environment [94]. Based on our findings, *"online learning"* (also called e-learning, virtual learning, virtual classroom, digital classroom, or distance learning) will help mitigate the impact of educational disruptions caused by the COVID-19 pandemic. While it may not be as effective as in-class learning in some cases, it will prevent potential brain drain that may result in the absence of continuous learning. Mobile and web-based learning platforms, many of which are available today, should be readily accessible in schools at all levels going forward. Designers should ensure these tools provide personalized learning experience such that students can manage their own content and the tools offer tailored suggestions that fit their interests or needs. The tools should also support collaborative learning where students can work together on assignments, projects, or other tasks similar to what they do in the real-world. Furthermore, governments across the globe should ensure equitable access to these educational technologies irrespective of economic, financial, racial, or cultural differences. Public infrastructure supporting these technologies (such as stable electricity, as well as affordable and reliable internet) should be considered a top priority and made readily accessible to people.

Finally, governments at all levels should partner (rather than compete) with health professionals and researchers to form a strong force against COVID-19. Our findings revealed *"advocacy for testing"* which reflects public call for increased testing since some governments are still struggling in this area due to their political interests superseding public health. Research argues the significance and effectiveness of adaptive evidence-making intervention (a fusion of scientific evidence and policy) during public health emergencies (such as the COVID-19 pandemic) [95]. This can only be possible if political leaders and health experts align and work harmoniously to address current and future pandemics.

## VI. Conclusion

We explored the impact of the COVID-19 pandemic on people globally using social media data. We analyzed over 1 million comments obtained from six social media platforms using a seven-stage context-aware Natural Language Processing (NLP) approach to extract relevant keyphrases which were further categorized into broader themes. Our results revealed 34 negative themes and 20 positive themes surrounding the COVID-19 pandemic. We discussed the economic, socio-political, educational, and political issues and suggested interventions to tackle them based on the positive themes and research evidence. These interventions would inform and help governments, organizations, and individuals to minimize the spread and impact of COVID-19 and to respond effectively to future pandemics.

## Acknowledgments

The authors would like to thank Dalhousie University's DeepSense team and Compute Canada for supporting this research with the computing infrastructure required for data analysis.

## References

[1] D. M. Morens, G. K. Folkers, and A. S. Fauci, "The challenge of emerging and re-emerging infectious diseases," *Nature*, vol. 430, no. 6996. Nature Publishing Group, pp. 242–249, 08-Jul-2004, doi: 10.1038/nature02759.

[2] T. N. Jilani and A. H. Siddiqui, *H1N1 Influenza (Swine Flu)*. StatPearls Publishing, 2018.

[3] F. S. Dawood *et al.*, "Estimated global mortality associated with the first 12 months of 2009 pandemic influenza A H1N1 virus circulation: A modelling study," *Lancet Infect. Dis.*, vol. 12, no. 9, pp. 687–695, Sep. 2012, doi: 10.1016/S1473-3099(12)70121-4.

[4] World Health Organization, "HIV/AIDS," 2019. [Online]. Available: https://www.who.int/news-room/fact-sheets/detail/hiv-aids. [Accessed: 16-May-2020].

[5] H. Tian *et al.*, "An investigation of transmission control measures during the first 50 days of the COVID-19 epidemic in China," *Science (80-. ).*, vol. 368, no. 6491, p. eabb6105, Mar. 2020, doi: 10.1126/science.abb6105.

[6] F. Wu *et al.*, "A new coronavirus associated with human respiratory disease in China," *Nature*, vol. 579, no. 7798, pp. 265–269, Mar. 2020, doi: 10.1038/s41586-020-2008-3.

[7] Johns Hopkins Coronavirus Resource Center, "COVID-19 Map." [Online]. Available: https://coronavirus.jhu.edu/map.html. [Accessed: 16-May-2020].

[8] K. E. Jones *et al.*, "Global trends in emerging infectious diseases," *Nature*, vol. 451, no. 7181, pp. 990–993, Feb. 2008, doi: 10.1038/nature06536.

[9] D. E. Bloom, S. Black, and R. Rappuoli, "Emerging infectious diseases: A proactive approach," *Proceedings of the National Academy of Sciences of the United States of America*, vol. 114, no. 16. National Academy of Sciences, pp. 4055–4059, 18-Apr-2017, doi: 10.1073/pnas.1701410114.

[10] V. Fan, D. Jamison, and L. Summers, "The Inclusive Cost of Pandemic Influenza Risk," *Natl. Bur. Econ. Res.*, 2016, doi: 10.3386/w22137.

[11] G. Barbier and H. Liu, "Data Mining in Social Media," in *Social*

*Network Data Analytics*, Springer US, 2011, pp. 327–352.
[12] S. Kemp, "Digital 2020: Global Digital Overview," 2020. [Online]. Available: https://datareportal.com/reports/digital-2020-global-digital-overview. [Accessed: 17-May-2020].
[13] P. Robinson, D. Turk, S. Jilka, and M. Cella, "Measuring attitudes towards mental health using social media: investigating stigma and trivialisation," *Soc. Psychiatry Psychiatr. Epidemiol.*, vol. 54, no. 1, pp. 51–58, Jan. 2019, doi: 10.1007/s00127-018-1571-5.
[14] S. C. Guntuku, A. Buffone, K. Jaidka, J. Eichstaedt, and L. Ungar, "Understanding and Measuring Psychological Stress using Social Media," in *Proceedings of the 13th International Conference on Web and Social Media, ICWSM 2019*, 2018, pp. 214–225.
[15] Y. Zhan, J.-F. Etter, S. Leischow, and D. Zeng, "Electronic cigarette usage patterns: a case study combining survey and social media data," *J. Am. Med. Informatics Assoc.*, vol. 26, no. 1, pp. 9–18, 2019, doi: 10.1093/jamia/ocy140.
[16] S. Hassanpour, N. Tomita, T. DeLise, B. Crosier, and L. A. Marsch, "Identifying substance use risk based on deep neural networks and Instagram social media data," *Neuropsychopharmacology*, vol. 44, no. 3, pp. 487–494, Feb. 2019, doi: 10.1038/s41386-018-0247-x.
[17] Y. Huang, D. Huang, and Q. C. Nguyen, "Census Tract Food Tweets and Chronic Disease Outcomes in the U.S., 2015–2018," *Int. J. Environ. Res. Public Health*, vol. 16, no. 6, p. 975, Mar. 2019, doi: 10.3390/ijerph16060975.
[18] O. Oyebode and R. Orji, "Detecting Factors Responsible for Diabetes Prevalence in Nigeria using Social Media and Machine Learning," in *15th International Conference on Network and Service Management, CNSM 2019*, 2019, doi: 10.23919/CNSM46954.2019.9012679.
[19] C. Chew and G. Eysenbach, "Pandemics in the age of Twitter: Content analysis of tweets during the 2009 H1N1 outbreak," *PLoS One*, vol. 5, no. 11, 2010, doi: 10.1371/journal.pone.0014118.
[20] A. Signorini, A. M. Segre, and P. M. Polgreen, "The Use of Twitter to Track Levels of Disease Activity and Public Concern in the U.S. during the Influenza A H1N1 Pandemic," *PLoS One*, vol. 6, no. 5, p. e19467, May 2011, doi: 10.1371/journal.pone.0019467.
[21] O. Oyebode and R. Orji, "Social Media and Sentiment Analysis: The Nigeria Presidential Election 2019," in *2019 IEEE 10th Annual Information Technology, Electronics and Mobile Communication Conference, IEMCON 2019*, 2019, pp. 140–146, doi: 10.1109/IEMCON.2019.8936139.
[22] A. Tumasjan., T. O. Sprenger, P. G. Sandner., and I. M. Welpe, "Predicting Elections with Twitter: What 140 Characters Reveal about Political Sentiment," *Proc. Fourth Int. AAAI Conf. Weblogs Soc. Media Predict.*, 2010, doi: 10.1074/jbc.M501708200.
[23] W. Budiharto and M. Meiliana, "Prediction and analysis of Indonesia Presidential election from Twitter using sentiment analysis," *J. Big Data*, vol. 5, no. 1, Dec. 2018, doi: 10.1186/s40537-018-0164-1.
[24] E. Tjong, K. Sang, and J. Bos, "Predicting the 2011 Dutch Senate Election Results with Twitter," *13th Conf. Eur. Chapter Assoc. Comput. Linguist.*, no. 53, pp. 65–72, 2012.
[25] J. Ma, Y. K. Tse, X. Wang, and M. Zhang, "Examining customer perception and behaviour through social media research – An empirical study of the United Airlines overbooking crisis," *Transp. Res. Part E Logist. Transp. Rev.*, vol. 127, pp. 192–205, Jul. 2019, doi: 10.1016/j.tre.2019.05.004.
[26] N. F. Ibrahim and X. Wang, "Decoding the sentiment dynamics of online retailing customers: Time series analysis of social media," *Comput. Human Behav.*, vol. 96, pp. 32–45, Jul. 2019, doi: 10.1016/j.chb.2019.02.004.
[27] H. C. Tissot *et al.*, "Natural language processing for mimicking clinical trial recruitment in critical care: a semi-automated simulation based on the LeoPARDS trial," *IEEE J. Biomed. Heal. Informatics*, pp. 1–1, Mar. 2020, doi: 10.1109/jbhi.2020.2977925.
[28] A. Vilic, J. A. Petersen, K. Hoppe, and H. B. D. Sorensen, "Visualizing patient journals by combining vital signs monitoring and natural language processing," in *Proceedings of the Annual International Conference of the IEEE Engineering in Medicine and Biology Society, EMBS*, 2016, vol. 2016-October, pp. 2529–2532, doi: 10.1109/EMBC.2016.7591245.
[29] F. J. Grajales, S. Sheps, K. Ho, H. Novak-Lauscher, and G. Eysenbach, "Social media: A review and tutorial of applications in medicine and health care," *J. Med. Internet Res.*, vol. 16, no. 2, pp. 1–68, Feb. 2014, doi: 10.2196/jmir.2912.
[30] O. Oyebode *et al.*, "Health, Psychosocial, and Social issues emanating from COVID-19 pandemic based on Social Media Comments using Natural Language Processing," Jul. 2020.
[31] A. Park and M. Conway, "Tracking Health Related Discussions on Reddit for Public Health Applications," *AMIA ... Annu. Symp. proceedings. AMIA Symp.*, vol. 2017, pp. 1362–1371, 2017.
[32] H. Jelodar, Y. Wang, R. Orji, and H. Huang, "Deep Sentiment Classification and Topic Discovery on Novel Coronavirus or COVID-19 Online Discussions: NLP Using LSTM Recurrent Neural Network Approach," *IEEE J. Biomed. Heal. Informatics*, pp. 1–1, Jun. 2020, doi: 10.1109/jbhi.2020.3001216.
[33] T. Bekhuis, M. Kreinacke, H. Spallek, M. Song, and J. A. O'Donnell, "Using natural language processing to enable in-depth analysis of clinical messages posted to an Internet mailing list: a feasibility study.," *J. Med. Internet Res.*, vol. 13, no. 4, p. e98, Nov. 2011, doi: 10.2196/jmir.1799.
[34] M. J. Paul and M. Dredze, "You Are What You Tweet: Analyzing Twitter for Public Health," Jul. 2011.
[35] A. L. Nobles, C. N. Dreisbach, J. Keim-Malpass, and L. E. Barnes, "'Is This an STD? Please Help!''' Online Information Seeking for Sexually Transmitted Diseases on Reddit,'" Jun. 2018.
[36] A. McNeill, P. R. Harris, and P. Briggs, "Twitter Influence on UK Vaccination and Antiviral Uptake during the 2009 H1N1 Pandemic," *Front. Public Heal.*, vol. 4, p. 26, Feb. 2016, doi: 10.3389/fpubh.2016.00026.
[37] A. Agarwal, C. Baechle, R. Behara, and X. Zhu, "A Natural Language Processing Framework for Assessing Hospital Readmissions for Patients with COPD," *IEEE J. Biomed. Heal. Informatics*, vol. 22, no. 2, pp. 588–596, Mar. 2018, doi: 10.1109/JBHI.2017.2684121.
[38] O. Oyebode, F. Alqahtani, and R. Orji, "Using Machine Learning and Thematic Analysis Methods to Evaluate Mental Health Apps Based on User Reviews," *IEEE Access*, pp. 1–1, Jun. 2020, doi: 10.1109/access.2020.3002176.
[39] K. Dave and V. Varma, "Pattern based keyword extraction for Contextual Advertising," in *International Conference on Information and Knowledge Management, Proceedings*, 2010, pp. 1885–1888, doi: 10.1145/1871437.1871754.
[40] Twitter Inc., "Consuming streaming data." [Online]. Available: https://developer.twitter.com/en/docs/tutorials/consuming-streaming-data. [Accessed: 20-May-2020].
[41] Google Inc., "YouTube Data API." [Online]. Available: https://developers.google.com/youtube/v3. [Accessed: 20-May-2020].
[42] Nlife Media, "Corona Virus Panic/Discussion Thread - General Discussion Forum." [Online]. Available: https://www.pushsquare.com/forums/ps_general_discussion/corona_virus_panicdiscussion_thread. [Accessed: 02-Apr-2020].
[43] Archinect, "corona virus covid-19 and you | Forum." [Online]. Available: https://archinect.com/forum/thread/150187455/corona-virus-covid-19-and-you. [Accessed: 02-Apr-2020].
[44] Archinect, "COVID - 19 Thread Central | Forum | Archinect." [Online]. Available: https://archinect.com/forum/thread/150188615/covid-19-thread-central. [Accessed: 02-Apr-2020].
[45] Future US Inc., "Coronavirus & Epidemiology | Live Science Forums." [Online]. Available: https://forums.livescience.com/forums/coronavirus-epidemiology.42/. [Accessed: 02-Apr-2020].
[46] "Slang Words Dictionary." [Online]. Available: https://raw.githubusercontent.com/sifei/Dictionary-for-Sentiment-Analysis/master/slang/acrynom.csv. [Accessed: 19-Jun-2019].
[47] "Slang Lookup Table." [Online]. Available: https://raw.githubusercontent.com/felipebravom/StaticTwitterSent/master/extra/SentiStrength/SlangLookupTable.txt. [Accessed: 19-Jun-2019].
[48] PyPI, "langdetect 1.0.8." [Online]. Available: https://pypi.org/project/langdetect/. [Accessed: 09-Aug-2020].
[49] E. F. Tjong Kim Sang and F. De Meulder, "Introduction to the CoNLL-2003 shared task," in *Proceedings of the seventh conference on Natural language learning at HLT-NAACL 2003 -*, 2003, vol. 4, pp. 142–147, doi: 10.3115/1119176.1119195.
[50] J. A. Asmuth and D. Gentner, "Context sensitivity of relational nouns," in *Proceedings of the 27th Annual Meeting of the Cognitive Science Society*, 2005, pp. 163–168.
[51] P. Chesley, B. Vincent, L. Xu, and R. K. Srihari, "Using verbs and adjectives to automatically classify blog sentiment," *Training*, vol. 580, no. 263, p. 233, 2006.
[52] B. Santorini, "Part-of-speech tagging guidelines for the penn treebank project (3rd revision)," *Tech. Reports*, p. 570, 1990.
[53] A. Taylor, M. Marcus, and B. Santorini, "The Penn Treebank: An Overview," in *Treebanks: Building and Using Parsed Corpora*,

Springer, Dordrecht, 2003, pp. 5–22.
[54] "nltk.tokenize package — NLTK 3.5 documentation." [Online]. Available: http://www.nltk.org/api/nltk.tokenize.html?highlight=tokenizer#nltk.tokenize.punkt.PunktSentenceTokenizer. [Accessed: 23-May-2020].
[55] P. Han, S. Shen, D. Wang, and Y. Liu, "The influence of word normalization in English document clustering," in *CSAE 2012 - Proceedings, 2012 IEEE International Conference on Computer Science and Automation Engineering*, 2012, vol. 2, pp. 116–120, doi: 10.1109/CSAE.2012.6272740.
[56] B. Liu, "Sentence Subjectivity and Sentiment Classification," in *Sentiment Analysis: Mining Opinions, Sentiments, and Emotions*, 2015, pp. 70–88.
[57] C. J. Hutto and E. Gilbert, "VADER: A Parsimonious Rule-based Model for Sentiment Analysis of Social Media Text," in *Eighth International AAAI Conference on Weblogs and Social Media*, 2014, pp. 216–225.
[58] D. L. Blustein, R. Duffy, J. A. Ferreira, V. Cohen-Scali, R. G. Cinamon, and B. A. Allan, "Unemployment in the time of COVID-19: A research agenda," *Journal of Vocational Behavior*, vol. 119. Academic Press Inc., p. 103436, 01-Jun-2020, doi: 10.1016/j.jvb.2020.103436.
[59] W. Kawohl and C. Nordt, "COVID-19, unemployment, and suicide," *The Lancet Psychiatry*, vol. 7, no. 5. Elsevier Ltd, pp. 389–390, 01-May-2020, doi: 10.1016/S2215-0366(20)30141-3.
[60] R. Fairlie, K. Couch, and H. Xu, "The Impacts of COVID-19 on Minority Unemployment: First Evidence from April 2020 CPS Microdata," Cambridge, MA, May 2020.
[61] O. Coibion, Y. Gorodnichenko, and M. Weber, "The Cost of the Covid-19 Crisis: Lockdowns, Macroeconomic Expectations, and Consumer Spending," Cambridge, MA, May 2020.
[62] A. W. Bartik, M. Bertrand, Z. B. Cullen, E. L. Glaeser, M. Luca, and C. T. Stanton, "How Are Small Businesses Adjusting to COVID-19? Early Evidence from a Survey," 2020.
[63] T. Didier, F. Huneeus, M. Larrain, and S. L. Schmukler, *Financing Firms in Hibernation during the COVID-19 Pandemic*. The World Bank, 2020.
[64] Constantino Hevia and Pablo Andrés Neumeyer, "A perfect storm: COVID-19 in emerging economies | VOX, CEPR Policy Portal," 2020. [Online]. Available: https://voxeu.org/article/perfect-storm-covid-19-emerging-economies. [Accessed: 21-Jun-2020].
[65] S. Baker, N. Bloom, S. Davis, K. Kost, M. Sammon, and T. Viratyosin, "The Unprecedented Stock Market Impact of COVID-19," *Natl. Bur. Econ. Res.*, 2020, doi: 10.3386/w26945.
[66] M. Nicola *et al.*, "The socio-economic implications of the coronavirus pandemic (COVID-19): A review," *International Journal of Surgery*, vol. 78. Elsevier Ltd, pp. 185–193, 01-Jun-2020, doi: 10.1016/j.ijsu.2020.04.018.
[67] M. L. Ranney, V. Griffeth, and A. K. Jha, "Critical supply shortages - The need for ventilators and personal protective equipment during the Covid-19 pandemic," *New England Journal of Medicine*, vol. 382, no. 18. Massachussetts Medical Society, p. E41, 30-Apr-2020, doi: 10.1056/NEJMp2006141.
[68] G. Iacobucci, "Covid-19: Lack of PPE in care homes is risking spread of virus, leaders warn," *BMJ*, vol. 368, p. m1280, Mar. 2020, doi: 10.1136/bmj.m1280.
[69] D. Nogee and A. Tomassoni, "Concise Communication: Covid-19 and the N95 Respirator Shortage: Closing the Gap," *Infect. Control Hosp. Epidemiol.*, pp. 1–1, 2020, doi: 10.1017/ice.2020.124.
[70] E. Goddard, "The impact of COVID-19 on food retail and food service in Canada: Preliminary assessment," *Can. J. Agric. Econ.*, 2020, doi: 10.1111/cjag.12243.
[71] Y. Wang *et al.*, "Impact of COVID-19 on blood centres in Zhejiang province China," *Vox Sang.*, p. vox.12931, Apr. 2020, doi: 10.1111/vox.12931.
[72] O. Akinleye, R. O. S. Dauda, O. Iwegub, and O. O. Popogbe, "Impact of COVID-19 Pandemic on Financial Health and Food Security: a Survey-Based Analysis," *SSRN Electron. J.*, Jun. 2020, doi: 10.2139/ssrn.3619245.
[73] T. Suzumura *et al.*, "The Impact of COVID-19 on Flight Networks," Jun. 2020.
[74] S. Villa *et al.*, "The COVID-19 pandemic preparedness ... or lack thereof: from China to Italy," *Glob. Heal. Med.*, vol. 2, no. 2, pp. 73–77, Apr. 2020, doi: 10.35772/ghm.2020.01016.
[75] K. Timmis and H. Brüssow, "The COVID-19 pandemic: some lessons learned about crisis preparedness and management, and the need for international benchmarking to reduce deficits," *Environ. Microbiol.*, Jun. 2020, doi: 10.1111/1462-2920.15029.
[76] K. Dolan *et al.*, "Global burden of HIV, viral hepatitis, and tuberculosis in prisoners and detainees," *The Lancet*, vol. 388, no. 10049. Lancet Publishing Group, pp. 1089–1102, 10-Sep-2016, doi: 10.1016/S0140-6736(16)30466-4.
[77] S. A. Kinner *et al.*, "Prisons and custodial settings are part of a comprehensive response to COVID-19," *The Lancet Public Health*, vol. 5, no. 4. Elsevier Ltd, pp. e188–e189, 01-Apr-2020, doi: 10.1016/S2468-2667(20)30058-X.
[78] W. Van Lancker and Z. Parolin, "COVID-19, school closures, and child poverty: a social crisis in the making," *The Lancet Public Health*, vol. 5, no. 5. Elsevier Ltd, pp. e243–e244, 01-May-2020, doi: 10.1016/S2468-2667(20)30084-0.
[79] A. Mian and S. Khan, "Coronavirus: The spread of misinformation," *BMC Medicine*, vol. 18, no. 1. BioMed Central Ltd., p. 89, 18-Dec-2020, doi: 10.1186/s12916-020-01556-3.
[80] D. A. Erku *et al.*, "When fear and misinformation go viral: Pharmacists' role in deterring medication misinformation during the 'infodemic' surrounding COVID-19," *Research in Social and Administrative Pharmacy*. Elsevier Inc., 01-May-2020, doi: 10.1016/j.sapharm.2020.04.032.
[81] V. A. Earnshaw and I. T. Katz, "Educate, Amplify, and Focus to Address COVID-19 Misinformation," *JAMA Heal. Forum*, vol. 1, no. 4, pp. e200460–e200460, Apr. 2020, doi: 10.1001/JAMAHEALTHFORUM.2020.0460.
[82] S. Laato, A. K. M. N. Islam, M. N. Islam, and E. Whelan, "Why do People Share Misinformation during the COVID-19 Pandemic?," Apr. 2020, doi: 10.1080/0960085X.2020.1770632.
[83] N. M. Nasir, B. Baequni, and M. I. Nurmansyah, "Misinformation related to COVID-19 in Indonesia," *J. Adm. Kesehat. Indones.*, vol. 8, no. 2, pp. 51–59, 2020.
[84] M. Motta, D. Stecula, and C. Farhart, "How Right-Leaning Media Coverage of COVID-19 Facilitated the Spread of Misinformation in the Early Stages of the Pandemic in the U.S.," *Can. J. Polit. Sci.*, pp. 1–8, 2020, doi: 10.1017/S0008423920000396.
[85] M. Painter and T. Qiu, "Political Beliefs affect Compliance with COVID-19 Social Distancing Orders," *SSRN Electron. J.*, Apr. 2020, doi: 10.2139/ssrn.3569098.
[86] G. Grossman, S. Kim, J. Rexer, and H. Thirumurthy, "Political Partisanship Influences Behavioral Responses to Governors' Recommendations for COVID-19 Prevention in the United States," *SSRN Electron. J.*, Apr. 2020, doi: 10.2139/ssrn.3578695.
[87] C. Adolph, K. Amano, B. Bang-Jensen, N. Fullman, and J. Wilkerson, "Pandemic Politics: Timing State-Level Social Distancing Responses to COVID-19," *medRxiv*, p. 2020.03.30.20046326, Mar. 2020, doi: 10.1101/2020.03.30.20046326.
[88] D. Stuckler, S. Basu, M. Suhrcke, A. Coutts, and M. McKee, "The public health effect of economic crises and alternative policy responses in Europe: an empirical analysis," *Lancet*, vol. 374, no. 9686, pp. 315–323, 2009, doi: 10.1016/S0140-6736(09)61124-7.
[89] U. Gentilini, M. Almenfi, I. Orton, and P. Dale, "Social Protection and Jobs Responses to COVID-19: A Real-Time Review of Country Measures," WB, Washington DC, Apr. 2020.
[90] S. E. Eikenberry *et al.*, "To mask or not to mask: Modeling the potential for face mask use by the general public to curtail the COVID-19 pandemic," *Infect. Dis. Model.*, vol. 5, pp. 293–308, Jan. 2020, doi: 10.1016/j.idm.2020.04.001.
[91] S. K. Paul and P. Chowdhury, "A production recovery plan in manufacturing supply chains for a high-demand item during COVID-19," *Int. J. Phys. Distrib. Logist. Manag.*, 2020, doi: 10.1108/IJPDLM-04-2020-0127.
[92] G. Gereffi, "What does the COVID-19 pandemic teach us about global value chains? The case of medical supplies," *J. Int. Bus. Policy*, pp. 1–15, Jul. 2020, doi: 10.1057/s42214-020-00062-w.
[93] Facebook Inc., "Coronavirus (COVID-19) Information Center." [Online]. Available: https://www.facebook.com/coronavirus_info. [Accessed: 13-Jul-2020].
[94] P. John and S. Wheeler, *The Digital Classroom: Harnessing technology for the future of learning and teaching*. Routledge, 2015.
[95] K. Lancaster, T. Rhodes, and M. Rosengarten, "Making evidence and policy in public health emergencies: lessons from COVID-19 for adaptive evidence-making and intervention," *Evid. Policy A J. Res. Debate Pract.*, Jun. 2020, doi: 10.1332/174426420x15913559981103.